%% file: Template.tex
\documentclass{article}
\usepackage{spconf,amsmath,graphicx}
\usepackage{enumitem}
\usepackage{amssymb,amsthm,enumerate,threeparttable,bm,subfigure,xcolor}
\usepackage{multirow}
\usepackage{caption}

\newcommand{\DACR}{\textbf{\texttt{DACR}}}
\newcommand{\DACL}{\textbf{\texttt{DACL}}}
\newcommand{\IDAR}{\textbf{\texttt{IDAR}}}
\def\blue#1{\textcolor{blue}{#1}}

\title{DACR: Distribution-Augmented Contrastive Reconstruction for Time-Series Anomaly Detection}
\name{Lixu Wang\textsuperscript{1}, Shichao Xu\textsuperscript{1}, Xinyu Du\textsuperscript{2}, Qi Zhu\textsuperscript{1}}
\address{\textsuperscript{1}Northwestern University, IL, USA\\
\textsuperscript{2}General Motors Global R\&D, MI, USA\thanks{We gratefully acknowledge the support from the NSF awards 1834701, 1724341, 2038853, and a grant from General Motors.}} 
\begin{document}
%\ninept
%
% 
\maketitle

\begin{abstract}
Anomaly detection in time-series data is crucial for identifying faults, failures, threats, and outliers across a range of applications. Recently, deep learning techniques have been applied to this topic, but they often struggle in real-world scenarios that are complex and highly dynamic, e.g., the normal data may consist of multiple distributions, and various types of anomalies may differ from the normal data to different degrees. In this work, to tackle these challenges, we propose \textit{Distribution-Augmented Contrastive Reconstruction} (\DACR{}). \DACR{} generates extra data disjoint from the normal data distribution to compress the normal data's representation space, and enhances the feature extractor through contrastive learning to better capture the intrinsic semantics from time-series data. Furthermore, \DACR{} employs an attention mechanism to model the semantic dependencies among multivariate time-series features, thereby achieving more robust reconstruction for anomaly detection. Extensive experiments conducted on nine benchmark datasets in various anomaly detection scenarios demonstrate the effectiveness of \DACR{} in achieving new state-of-the-art time-series anomaly detection.
\end{abstract}

\input{introduction}

\input{relatedwork}

\input{method}

\input{experiment}
\input{conclusion}

\vfill\pagebreak

\bibliographystyle{IEEEbib}
\bibliography{refs}

\end{document}

%% file: introduction.tex
\section{Introduction}
System malfunctions and anomalies are unavoidable in many real-world applications across various fields~\cite{blazquez2021review}. Accurate anomaly detection is critically important for monitoring and alarming potential faults, threats, and risks in these systems~\cite{chen2022tcn}. Recently, data-driven methods have become the mainstream for anomaly detection, among which algorithms based on deep learning perform the best and can be divided into three categories: reconstruction prediction~\cite{su2019robust,li2021multivariate}, anomaly exposure~\cite{hendrycks2018deep,goyal2020drocc}, and self-supervised learning (SSL)~\cite{tack2020csi,qiu2021neural}. Reconstruction prediction utilizes the difference between the reconstruction outputs for normal and abnormal data. Anomaly exposure synthesizes extra data that is different from the normal data to better model the normal data distribution. SSL leverages auxiliary tasks to help models extract the intrinsic semantics of normal data.

\begin{figure*}[h]
\centering
\includegraphics[width=1.\textwidth]{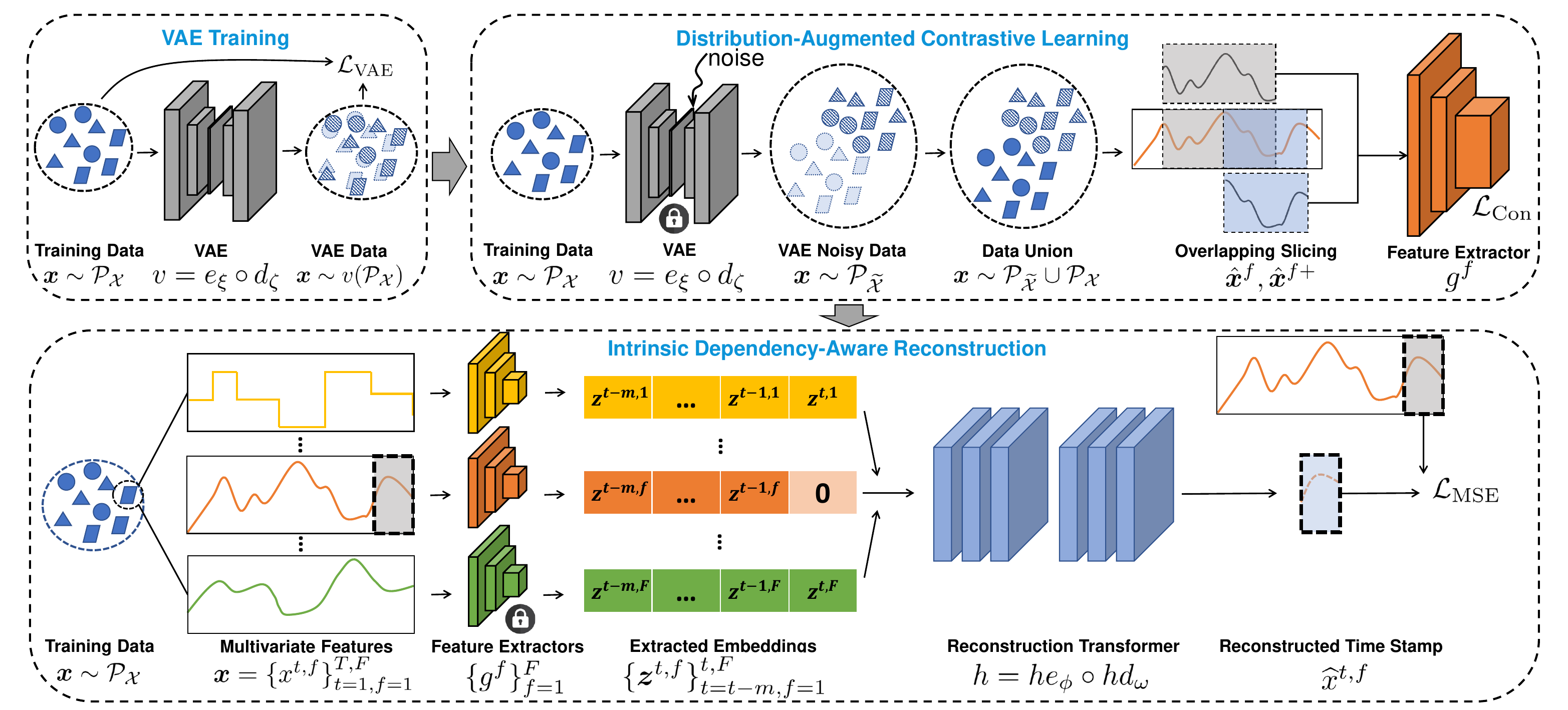}
\vspace{-20pt}
\caption{Overall workflow of \emph{Distribution-Augmented Contrastive Reconstruction} (\DACR{}) for time-series anomaly detection.}
\vspace{-12pt}
\label{Fig_overview}
\end{figure*}
 
However, anomaly detection scenarios in the real world are often very complex and highly dynamic. For example, normal data may consist of multiple distributions, and various types of anomalies may differ from normal data to different degrees. When faced with these challenging scenarios, the aforementioned deep anomaly detection methods expose a number of shortcomings. For instance, reconstruction prediction~\cite{su2019robust,li2021multivariate} performs poorly when the normal data consists of multiple distributions that have different learning difficulties. The performance of anomaly exposure~\cite{hendrycks2018deep,goyal2020drocc} depends on the similarity between ground-truth anomaly data and simulated extra data. SSL~\cite{tack2020csi,qiu2021neural} performs better when dealing with the above scenarios, but has its own challenges. For example, the feature extractor trained with contrastive learning~\cite{hjelm2018learning} eventually converges to a uniform hyperspherical space~\cite{wang2020understanding} that is not suitable for anomaly detection~\cite{sohn2020learning}.

To address the aforementioned challenges, we propose a novel method called  \emph{Distribution-Augmented Contrastive Reconstruction} (\DACR{}), which comprises three stages. Specifically, in the first stage, we train a variational auto-encoder (VAE) to reconstruct normal data. In the second stage, we introduce random noise into the latent space when applying the trained VAE to generate extra data from a different distribution, a process we refer to as \emph{distribution augmentation}. With the extra data, a series of simple feature extractors are trained with contrastive learning, enabling them to extract intrinsic semantics from each univariate time-series feature. In the final stage, \DACR{} employs a transformer to model the inter-feature semantic dependency. \textbf{This allows \DACR{} to reconstruct time series on the basis of intrinsic semantics rather than overfitting to artificial features that are only specific to the reconstruction task, with such being highly generalizable to more anomalies}. The overall workflow of \DACR{} is depicted in Fig.~\ref{Fig_overview}. Extensive experiments on nine benchmark datasets, which encompass various anomaly detection scenarios, demonstrate that our methods outperform existing state-of-the-art baseline methods substantially. In summary, the main contributions include: 

1. We address practical anomaly detection scenarios where the normal data or the anomalies consist of multiple distributions. The challenge is to model the implicit discrepancies between these normal and anomaly distributions.

2. We develop \DACR{}, which combines the strengths of self-supervised learning and attention-based reconstruction prediction. By capturing the intrinsic semantic dependency between multivariate features of time series, \DACR{} is significantly more sensitive to potential anomalies.

3. We conduct extensive experiments on nine benchmark datasets, covering scenarios with varying degrees of discrepancies between normal and anomalous data. Experiment results demonstrate that \DACR{} consistently and substantially outperforms existing state-of-the-art baselines.

%% file: relatedwork.tex
\smallskip
\noindent \textbf{Related Work:}
For \emph{reconstruction prediction}, anomaly detectors are strongly over-fitted to normal data. Recent studies apply convolution neural networks~\cite{chauhan2022robust}, LSTM~\cite{malhotra2016lstm}, and transformers~\cite{tuli2022tranad} to build such reconstruction models. \emph{Anomaly exposure} usually relies on extra data generation, and by training the model in supervised learning~\cite{hendrycks2018deep, goyal2020drocc, wang2021non} to distinguish extra data from normal data, the distribution of normal data can be better modeled. \emph{SSL-based methods} rely on various auxiliary tasks, such as rotation prediction~\cite{komodakis2018unsupervised} and data augmentation~\cite{sohn2020learning, tack2020csi, yue2022ts2vec, qiu2021neural, reiss2021mean, mohapatra2023efficient}. While these methods are designed for image data, no study explores whether SSL can work on time-series anomaly detection. Other anomaly detection works include graph neural network~\cite{deng2021graph} and ensemble learning~\cite{ums}.

%% file: method.tex
\section{Methodology}
\textbf{Problem Formulation:} Suppose an unlabeled dataset with $N$ time-series samples $\mathcal{X} \!=\! \{{\bm x}_i \| {\bm x}_i \sim \mathcal{P}_{\mathcal{X}}\}_{i=1}^N$ is given, where $\mathcal{P}_{\mathcal{X}}$ is the input feature distribution. Each sample ${\bm x}_i$ can be regarded as an observation of a matrix-valued random variable with dimensions $T \!\times\! F$, where $T$ is the sequence length and $F$ is the feature dimension, i.e., ${\bm x}_i=\{x_i^{t, f}\}_{t=1, f=1}^{T, F}$. Following standard anomaly detection assumptions~\cite{standardad, gdn}, dataset $\mathcal{X}$ is considered as being full of normal data. The objective is to learn a model based on $\mathcal{X}$ to accurately infer each time stamp of each testing sample as either normal or anomaly.

\subsection{Distribution-Augmented Contrastive Learning}
\label{sec:dacl}

\noindent \textbf{Time Series Contrastive Learning}

Different from contrastive learning on visual data~\cite{sohn2020learning,tack2020csi}, it is not yet straightforward to find suitable augmentation techniques for producing positive pairs of time-series data. Inspired by TS2Vec~\cite{yue2022ts2vec}, we extend overlapping slicing to our problem. More specifically, overlapping slicing here means randomly cutting out two fragments ($\hat{\bm x}_i, \hat{\bm x}_i^+$) from a given time-series instance ${\bm x}_i$ while ensuring that there is an overlapping part between them, i.e., $\hat{\bm x}_i \!=\! \{x_i^{t}\}_{t=a}^{b}$, $\hat{\bm x}_i^+ \!=\! \{x_i^{t}\}_{t=c}^{d}$, where $1 \!\leq\! a\!<\!c\!<\!b\!<\!d \!\leq\! T$. The instance-wise contrastive comparison between positive pairs ($\hat{\bm z}_i^t \,\&\, \hat{\bm z}_i^{t+}$) is constructed as comparing representations of fragments from the same data instance, while that of negative pairs ($\hat{\bm z}_i^t \,\&\, \hat{\bm z}_j^{t+}$, and $\hat{\bm z}_i^t \,\&\, \hat{\bm z}_j^{t}$ where $i \neq j$) is constructed as comparing representations of fragments from different data instances.
\vspace{-5pt}
\begin{equation}
\mathcal{L}_{\mathrm{In}, i}^t = -\log \frac{\exp(\hat{\bm z}_i^t \cdot \hat{\bm z}_i^{t+})}{\sum_j\left(\exp(\hat{\bm z}_i^t \!\cdot\! \hat{\bm z}_j^{t+}) + \mathbf{1}_{i \neq j}\exp(\hat{\bm z}_i^t \cdot \hat{\bm z}_j^{t})\right)}.
\vspace{-5pt}
\label{Eq_instance_contrastive}
\end{equation}
Here the range of $j$ is $[1, N_B]$ if the batch size is $N_B$. $\mathbf{1}_{(\cdot)}$ is an indicator function so that if the subscript condition is true, $\mathbf{1}_{\mathrm{True}}\!=\!1$, otherwise, $\mathbf{1}_{\mathrm{False}}\!=\!0$. Considering the temporal consistency in time-series data, we also need to conduct temporal contrastive comparisons. However, we only consider the comparison in the overlapping part $t \!\in\! [c, b]$ of the augmented fragments $\hat{\bm x}_i, \hat{\bm x}_i^+$, instead of considering the entire sequence as in TS2Vec~\cite{yue2022ts2vec}. The positive pairs ($\hat{\bm z}_i^t \,\&\, \hat{\bm z}_i^{t+}$) of temporal CL are constructed as the representations of augmented fragments at the same time stamp, while the negative pairs ($\hat{\bm z}_i^t \,\&\, \hat{\bm z}_i^{t^\prime +}$, and $\hat{\bm z}_i^t \,\&\, \hat{\bm z}_i^{t^\prime}$ where $t \neq t^\prime$) are the representations of augmented fragments at different time stamps.
\vspace{-5pt}
\begin{equation}
\vspace{-5pt}
\mathcal{L}_{\mathrm{Te}, i}^t = -\log \frac{\exp(\hat{\bm z}_i^t \cdot \hat{\bm z}_i^{t+})}{\sum_{t^\prime}\left(\exp(\hat{\bm z}_i^t \!\cdot\! \hat{\bm z}_i^{t^\prime +}) + \mathbf{1}_{t \neq t^\prime}\exp(\hat{\bm z}_i^t \cdot \hat{\bm z}_i^{t^\prime})\right)}.
\end{equation}
Finally, for a data batch in the mini-batch training, we have an overall contrastive loss as:
\vspace{-5pt}
\begin{equation}
\vspace{-5pt}
\mathcal{L}_{\mathrm{Con}} = \frac{1}{N_B (b-c)}\sum_{i=1}^{N_B} \sum_{t=c}^b \left(\mathcal{L}_{\mathrm{In}, i}^t + \mathcal{L}_{\mathrm{Te}, i}^t\right).
\label{Eq_contrastive_loss}
\end{equation}

\noindent \textbf{VAE-Based Distribution Augmentation}

For standard CL as shown in Eq.~\eqref{Eq_instance_contrastive}, it has been shown that the optimal solution shapes like a perfect uniform distribution for all training data in the representation space~\cite{wang2020understanding}. In such cases, it is difficult to distinguish outliers from their proximal inliers (training data). In this work, we generate extra data from a disjoint distribution to the normal data to occupy a certain space of the final uniform distribution. Through this \emph{distribution augmentation} process, the uniformity of the original normal data is greatly reduced. To make the extra data diverse enough~\cite{liu2022deja}, we achieve that by introducing random noise to a variational auto-encoder (VAE) $v=e_\xi \circ d_\zeta$. Specifically, we first train $v$ with the task of reconstructing the input data. The training loss is Mean Square Error (MSE) and Kullback–Leibler Divergence, as shown below:
\vspace{-5pt}
\begin{equation}
\vspace{-5pt}
\mathcal{L}_{\mathrm{VAE}} = \mathbb{E}_{{\bm x}_i \sim \mathcal{P}_{\mathcal{X}}}\|{\bm x}_i - v({\bm x}_i)\|_2 + \mathcal{D}_{\mathrm{KL}}(\mathcal{P}_{\mathcal{X}} \| v(\mathcal{P}_{\mathcal{X}})).
\label{Eq_VAE_loss}
\end{equation}
After the VAE training, we can generate extra data from disjoint distributions with various levels of discrepancies by injecting Gaussian noise into the low-dimensional latent space:
\vspace{-5pt}
\begin{equation}
\vspace{-5pt}
e_\xi({\bm x}_i)^\prime = \alpha \odot e_\xi({\bm x}_i) + \beta,
\label{Eq_noise_injection}
\end{equation}
where $\alpha$ and $\beta$ are vectors of the same dimensions with $e_\xi({\bm x}_i)$, and we set $\alpha \sim \mathcal{N}(1, 0.1 \cdot \mathbf{I})$ and $\beta \sim \mathcal{N}(0, 0.1 \cdot \mathbf{I})$, where $\mathbf{I}$ is the unit matrix. We can use VAE decoder $d_\zeta$ to decode $e_\xi({\bm x}_i)^\prime$ into the input space, and then regard the decoded data as $\widetilde{\bm x}_i$. With the united dataset $\mathcal{X} \cup \widetilde{\mathcal{X}}$, we train a dedicated feature extractor $g^f$ for each univariate feature $f$ with the time-series contrastive loss (Eq.~\eqref{Eq_contrastive_loss}).

\subsection{Intrinsic Dependency-Aware Reconstruction}
\label{sec:idar}

As shown in Fig.~\ref{Fig_overview}, after the \DACL{} stage, we can obtain a feature extractor $g^f$ for each feature dimension $f$. In the third stage of our method, with Intrinsic Dependency-Aware Reconstruction (\IDAR{}) training, these feature extractors are all frozen, and a transformer model $h$ is incorporated to take the embedding vectors ${\bm z}$ produced by feature extractors as input and tries to reconstruct the time-series instances. 

Specifically, for example, suppose that our task is to reconstruct the $t$-th time stamp of the $f$-th feature dimension of the $i$-th time-series instance. The input matrix $I$ of $h$ is
\vspace{-5pt}
\begin{equation}
\vspace{-5pt}
\footnotesize
\left[[{\bm z}^{t-m, 1}, ..., {\bm z}^{t, 1}], ..., [{\bm z}^{t-m, f}, ..., {\bm z}^{t-1, f}, {\bm 0}], ..., [{\bm z}^{t-m, F}, ..., {\bm z}^{t, F}]\right],
\end{equation}
where $m$ is a hyper-parameter that controls how long the model can observe in history ($m\!=\!20$; please see Section~\ref{Sec_ablation_sensitivity} for the sensitivity analysis). Note that different from any autoregressive forecasting model, in addition to feeding historical time stamps of all feature dimensions, we also feed embeddings of the $t$-th stamp of all dimensions except for the $f$-th. The additional input can help the model better capture the inter-feature dependency of a shorter time period, improving the model's sensitivity to the anomalies. The used transformer architecture consists of an encoder $he_\phi$ and a decoder $hd_\omega$. To train $he_\phi$ and $hd_\omega$, we leverage the MSE loss as follows:
\vspace{-5pt}
\begin{equation}
\vspace{-5pt}
\mathcal{L}_{\mathrm{MSE}} = \frac{1}{N_BF(T-m)}\sum_{i=1}^{N_B} \sum_{t=m+1}^T \sum_{f=1}^{F} \|\widehat{x}_i^{t, f} - x_i^{t, f}\|_2^2.
\end{equation}
\textbf{Anomaly Scoring.} We also design an anomaly-scoring mechanism to coordinate the usage of \DACR{}, which picks the maximum increase percentage among all feature dimensions at the same time stamp compared to the maximum MSE error in the training data as the anomaly score:
\vspace{-5pt}
\begin{equation}
\vspace{-5pt}
S_i^t = \max_f \left\{\frac{\|\widehat{x}_i^{t, f} - x_i^{t, f}\|_2^2 - \mathrm{Err}^f}{\mathrm{Err}^f} \times 100\%\right\},
\end{equation}
where $\mathrm{Err}^f$ is the maximum MSE error of the $f$-th feature dimension among all training data $\mathcal{X}$. Finally, for a particular timestamp $t$ of a time-series instance ${\bm x}_i$, if its anomaly score $S_i^f$ is larger than zero, it is labeled as an anomaly.

%% file: experiment.tex
\section{Experimental Results}
\subsection{Experimental Settings}
Our code is implemented in PyTorch. All experiments are conducted on Ubuntu 18.04 LTS with NVIDIA TITAN RTX.

\begin{table}[h]
\centering
\caption{Time-Series Anomaly Detection Dataset Information (SAD). $N_{\text{sample}}$ -- sample quantity, $F$ -- dimension number, $N_C$ -- class number, and $T$ -- sequence length.}
\vspace{-10pt}
\resizebox{.5\textwidth}{!}{
\setlength{\tabcolsep}{0.85mm}{
\begin{tabular}{l|ccccc|cccc}
\hline
Dataset & SAD\cite{qiu2021neural} & CT & RS & PCoE\cite{saha2007nasa} & NAT & ASD\cite{xu2022calibrated} & SWaT & WaQ & SMD \\ \hline
$N_{\text{sample}}$ &8800  &2858  &300  &7565  &360  &8528  &0.48M  &0.14M  &28703  \\
$F$ &13  &3  &6  &3  &24  &19  &51  &11  &38  \\
$N_C$ &10  &20  &4  &4  &6  &-  &-  &-  &-  \\
$T$ &50  &182  &30  &256  &51  &100  &100  &100  &100  \\ \hline
\end{tabular}
}
}
\label{tab:dataset}
\vspace{-5pt}
\end{table}

\begin{figure*}
\centering
\begin{minipage}[h]{.6\textwidth}
\captionof{table}{Performance comparison between \DACR{} and baselines in the settings of both EAD (n=$N_C-$1) and IAD. AUC\textsubscript{$\pm$standard deviation} is used to evaluate the performance. \DACR{} significantly outperforms the second-best by 2.8-8.2\% on EAD, and 0.4-3.3\% on IAD.}
\vspace{-10pt}
\centering
\resizebox{1.\textwidth}{!}{
\setlength{\tabcolsep}{0.45mm}{
\begin{tabular}{l|ccccc|cccc}
\hline
\multirow{2}{*}{Baseline} & \multicolumn{5}{c|}{EAD} & \multicolumn{4}{c}{IAD} \\
 & SAD & CT & RS & PCoE & NAT & ASD & SWaT & WaQ & SMD \\ \hline
DROCC & 58.8\textsubscript{$\pm$0.5} & 57.6\textsubscript{$\pm$1.5} & 60.9\textsubscript{$\pm$0.2} & 69.7\textsubscript{$\pm$1.1} & 60.7\textsubscript{$\pm$1.6} & 69.3\textsubscript{$\pm$1.9} & 83.3\textsubscript{$\pm$0.9} & 68.9\textsubscript{$\pm$3.4} & 86.6\textsubscript{$\pm$0.9} \\
TS2Vec & 62.7\textsubscript{$\pm$0.6} & 62.4\textsubscript{$\pm$1.0} & 67.6\textsubscript{$\pm$0.6} & 72.4\textsubscript{$\pm$1.5} & 66.6\textsubscript{$\pm$1.8} & 78.9\textsubscript{$\pm$2.3} & 86.0\textsubscript{$\pm$1.8} & 67.8\textsubscript{$\pm$3.5} & 89.9\textsubscript{$\pm$0.9} \\
DROC & 63.5\textsubscript{$\pm$0.7} & 63.0\textsubscript{$\pm$0.7} & 68.9\textsubscript{$\pm$0.9} & 75.0\textsubscript{$\pm$1.1} & 69.2\textsubscript{$\pm$1.3} & 75.1\textsubscript{$\pm$1.5} & 84.9\textsubscript{$\pm$1.1} & 68.4\textsubscript{$\pm$2.2} & 86.6\textsubscript{$\pm$0.8} \\
MSC & 55.7\textsubscript{$\pm$2.0} & 58.0\textsubscript{$\pm$1.5} & 61.4\textsubscript{$\pm$1.0} & 65.0\textsubscript{$\pm$0.7} & 61.5\textsubscript{$\pm$0.9} & 80.0\textsubscript{$\pm$2.8} & 85.4\textsubscript{$\pm$0.9} & 70.6\textsubscript{$\pm$0.8} & 90.2\textsubscript{$\pm$1.3} \\
NTL & \textbf{85.1}\textsubscript{$\pm$0.3} & \textbf{87.4}\textsubscript{$\pm$0.2} & \textbf{80.0}\textsubscript{$\pm$0.4} & 75.5\textsubscript{$\pm$1.1} & \textbf{74.8}\textsubscript{$\pm$0.9} & 59.2\textsubscript{$\pm$4.5} & 85.0\textsubscript{$\pm$2.6} & 61.6\textsubscript{$\pm$9.1} & 74.6\textsubscript{$\pm$6.7} \\
GDN & 74.9\textsubscript{$\pm$2.1} & 66.4\textsubscript{$\pm$0.7} & 69.6\textsubscript{$\pm$0.9} & 73.8\textsubscript{$\pm$2.5} & 71.1\textsubscript{$\pm$1.3} & 77.9\textsubscript{$\pm$4.2} & 88.5\textsubscript{$\pm$3.6} & 65.9\textsubscript{$\pm$4.3} & 95.9\textsubscript{$\pm$1.6} \\
TranAD & 64.4\textsubscript{$\pm$1.1} & 61.3\textsubscript{$\pm$0.9} & 70.9\textsubscript{$\pm$0.6} & 72.7\textsubscript{$\pm$0.5} & 66.0\textsubscript{$\pm$1.0} & 91.5\textsubscript{$\pm$1.8} & 81.0\textsubscript{$\pm$0.7} & \textbf{72.9}\textsubscript{$\pm$1.7} & 66.2\textsubscript{$\pm$0.3} \\
COUTA & 65.0\textsubscript{$\pm$1.1} & 65.5\textsubscript{$\pm$0.8} & 72.2\textsubscript{$\pm$0.2} & 75.0\textsubscript{$\pm$0.9} & 68.0\textsubscript{$\pm$1.1} & \textbf{95.5}\textsubscript{$\pm$3.0} & \textbf{90.0}\textsubscript{$\pm$1.7} & 71.4\textsubscript{$\pm$0.6} & \textbf{98.4}\textsubscript{$\pm$1.5} \\
UMS & 68.0\textsubscript{$\pm$3.0} & 71.5\textsubscript{$\pm$0.6} & 75.2\textsubscript{$\pm$0.9} & 77.3\textsubscript{$\pm$1.4} & 69.7\textsubscript{$\pm$2.1} & 91.0\textsubscript{$\pm$2.5} & 86.6\textsubscript{$\pm$2.0} & 69.9\textsubscript{$\pm$1.6} & 96.5\textsubscript{$\pm$1.2} \\ \hline
\DACR{}-ab1 & 68.3\textsubscript{$\pm$3.3} & 82.0\textsubscript{$\pm$0.9} & 75.9\textsubscript{$\pm$0.5} & \textbf{78.5}\textsubscript{$\pm$0.9} & 69.8\textsubscript{$\pm$2.0} & 82.0\textsubscript{$\pm$1.1} & 87.1\textsubscript{$\pm$0.9} & 68.2\textsubscript{$\pm$1.3} & 87.0\textsubscript{$\pm$2.2} \\
\DACR{}-ab2 & 75.0\textsubscript{$\pm$2.2} & 66.9\textsubscript{$\pm$3.8} & 71.0\textsubscript{$\pm$0.7} & 74.5\textsubscript{$\pm$1.9} & 75.1\textsubscript{$\pm$1.0} & 88.0\textsubscript{$\pm$2.5} & 86.4\textsubscript{$\pm$0.8} & 72.0\textsubscript{$\pm$1.9} & 94.5\textsubscript{$\pm$0.6} \\
\DACR{} & \blue{\textbf{90.7}}\textsubscript{$\pm$0.3} & \blue{\textbf{91.9}}\textsubscript{$\pm$1.5} & \blue{\textbf{88.2}}\textsubscript{$\pm$3.2} & \blue{\textbf{81.3}}\textsubscript{$\pm$0.7} & \blue{\textbf{81.5}}\textsubscript{$\pm$1.6} & \blue{\textbf{96.2}}\textsubscript{$\pm$2.1} & \blue{\textbf{93.3}}\textsubscript{$\pm$0.9} & \blue{\textbf{75.9}}\textsubscript{$\pm$0.9} & \blue{\textbf{98.8}}\textsubscript{$\pm$1.0} \\ \hline
\end{tabular}
\label{tab:ad}
}
}
\end{minipage}
\begin{minipage}[h]{0.38\textwidth}
\centering
\includegraphics[width=1.\textwidth]{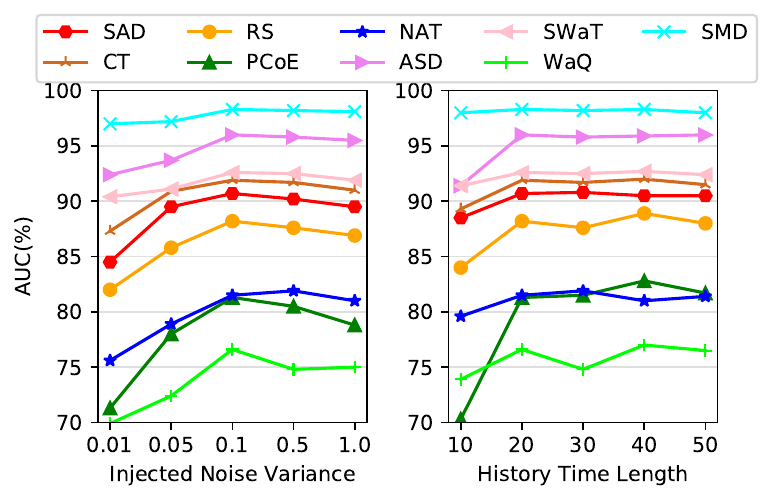}
% \vspace{-9pt}
\caption{Sensitivity analysis of various VAE injected noise degrees and different history time lengths of reconstruction transformer input.} 
\label{fig:sensitivity}
\end{minipage}
\vspace{-10pt}
\end{figure*}

\textbf{Dataset Settings.}
Two scenarios called Explicit Anomaly Detection (EAD) and Implicit Anomaly Detection (IAD) are considered. EAD refers to the setting where there is an explicit discrepancy between normal and abnormal data. We build the EAD by randomly selecting n classes as the normal data ($1\!<\!\text{n}\!<\!N_C$) and viewing the remaining classes as anomalies~\cite{qiu2021neural}. IAD refers to the setting where there is no explicit discrepancy, which is more challenging, since all normal data may have corresponding near out-of-distribution data. We follow COUTA~\cite{xu2022calibrated} to build the IAD setting. Please refer to Table~\ref{tab:dataset} for dataset details.

\textbf{Implementation Details.}
The batch size is set as 8. The learning rate of the Adam optimizer for training the feature extractor is 0.001, while that for VAE is 0.0001. For the EAD datasets, the default number of training iterations for the feature extractor is 200. For the IAD datasets, given the larger data size, we train 1000 iterations. We follow NTL~\cite{qiu2021neural} to conduct EAD based on the entire sequence of time-series instances and follow COUTA~\cite{xu2022calibrated} to adopt a window sliding mode for IAD. The backbone architecture of feature extractors is a dilated CNN~\cite{yue2022ts2vec}, and the VAE consists of 4 LSTM layers. We repeat our experiments 3 times with different seeds and report the average value and standard deviation for AUC. 

\textbf{Baseline Methods for Comparison.}
We compare \DACR{} with DROCC~\cite{goyal2020drocc}, TS2Vec~\cite{yue2022ts2vec}, DROC~\cite{sohn2020learning}, MSC~\cite{reiss2021mean}, NTL~\cite{qiu2021neural}, GDN~\cite{gdn}, TranAD~\cite{tuli2022tranad}, COUTA~\cite{xu2022calibrated}, UMS~\cite{ums}.

\subsection{Performance Comparison}
\begin{figure}[htbp]
\centering
\includegraphics[width=.4\textwidth]{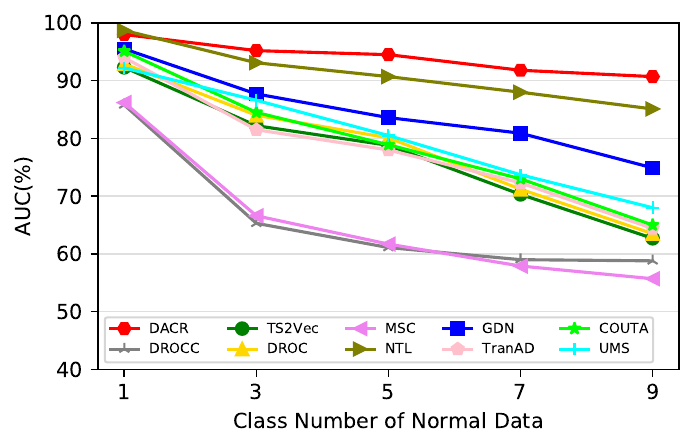}
\vspace{-12pt}
\caption{Performance comparison in EAD with different normal class numbers on the SAD dataset.}
\vspace{-5pt}
\label{fig:multiple_n_SAD}
\end{figure}

Table~\ref{tab:ad} presents experiment results of both EAD and IAD. From the results, we can see that \textbf{\DACR{} achieves the best performance for all cases in EAD and IAD, clearly outperforming baselines}. In addition, we also carry out experiments in setting different normal class numbers for EAD, and the results of the SAD dataset are shown in Fig.~\ref{fig:multiple_n_SAD}. The figure shows that \DACR{} always performs the best, demonstrating its effectiveness under challenging scenarios where the normal and abnormal data consist of multiple different distributions. The results on other datasets show similar trends.

\subsection{Ablation Study and Sensitivity Analysis}
\label{Sec_ablation_sensitivity}
We conducted an ablation study by modifying \DACR{} in two ways: 1) removing the VAE-based Distribution Augmentation (\DACR{}-ab1), or 2) replacing the \DACL{} feature extractor with that trained by GDN~\cite{gdn} (\DACR{}-ab2). The results in Table~\ref{tab:ad} show that the performance of \DACR{}-ab1 and \DACR{}-ab2 are significantly worse than \DACR{}, showing that all modules in \DACR{} are essential and complement each other well.
We conducted a sensitivity analysis on the noise degree ($\alpha, \beta$ in Eq.~\eqref{Eq_noise_injection}) with different variances (0.01, 0.05, 0.1, 0.5, 1.0). According to Fig.~\ref{fig:sensitivity}, we can observe that our method is not very sensitive to noise degrees larger than 0.05. The performance at 0.01 is poor because VAE cannot generate sufficiently diverse data at that moment. We also conducted a sensitivity analysis on the history time length of the transformer input, setting it from 10 to 50 with a stride of 10. Fig.~\ref{fig:sensitivity} shows that \DACR{} performs stably when the length exceeds 20.

%% file: conclusion.tex
\section{Conclusion}
We present \textit{Distribution-Augmented Contrastive Reconstruction} (\DACR{}) for time-series anomaly detection. \DACR{} leverages a VAE to conduct distribution augmentation, which helps extract intrinsic semantics from univariate time-series features through contrastive learning. Then \DACR{} applies the attention mechanism to model the semantic dependency between multivariate features and achieve reconstruction-based anomaly detection. Extensive experiments on nine benchmark datasets in various scenarios demonstrate that \DACR{} achieves new state-of-the-art performance.